\documentclass[conference]{IEEEtran}
\IEEEoverridecommandlockouts
\usepackage{cite}
\usepackage{amsmath,amssymb,amsfonts}
\usepackage{algorithmic}
\usepackage{graphicx}
\usepackage{textcomp}
\usepackage{xspace}
\usepackage{multirow}
\usepackage{xcolor}
\usepackage{hyperref}
\hypersetup{hidelinks,
	colorlinks=false,
	allcolors=blue,
	pdfstartview=Fit,
	breaklinks=true}

\def\BibTeX{{\rm B\kern-.05em{\sc i\kern-.025em b}\kern-.08em
    T\kern-.1667em\lower.7ex\hbox{E}\kern-.125emX}}
\begin{document}

\newcommand\system{DDT\xspace}
\newcommand\etal{{\it et~al.}}
\newcommand\ie{{\it i.e.}}
\newcommand\etc{{\it etc.}}
\newcommand\eg{{\it e.g.}}

\newcommand{\pink}[1]{\textcolor{magenta}{#1}}

\title{DDT: Dual-branch Deformable Transformer for Image Denoising}

\author{\IEEEauthorblockN{Kangliang Liu$^{1}$, Xiangcheng Du$^{1,2}$, Sijie Liu$^{1}$, Yingbin Zheng$^{2}$, Xingjiao Wu$^{1}$, Cheng Jin$^{1*}$\thanks{$^{*}$Corresponding author.}}
    \IEEEauthorblockA{$^1$\textit{School of Computer Science, Fudan University, Shanghai, China}~~$^2$\textit{Videt Technology, Shanghai, China}\\
    \{klliu21, xcdu22, sjliu22\}@m.fudan.edu.cn, zyb@videt.cn, \{xjwu\_cs, jc\}@fudan.edu.cn
    }    
}

\maketitle

\begin{abstract}
Transformer is beneficial for image denoising tasks since it can model long-range dependencies to overcome the limitations presented by inductive convolutional biases. However, directly applying the transformer structure to remove noise is challenging because its complexity grows quadratically with the spatial resolution. In this paper, we propose an efficient Dual-branch Deformable Transformer (\system) denoising network which captures both local and global interactions in parallel. We divide features with a fixed patch size and a fixed number of patches in local and global branches, respectively. In addition, we apply deformable attention operation in both branches, which helps the network focus on more important regions and further reduces computational complexity. We conduct extensive experiments on real-world and synthetic denoising tasks, and the proposed \system achieves state-of-the-art performance with significantly fewer computational costs. 
The source code and trained models are available at \pink{\href{https://github.com/Merenguelkl/DDT}{https://github.com/Merenguelkl/DDT}}.
\end{abstract}

\begin{IEEEkeywords}
Image denoising, dual-branch, transformer, deformable attention
\end{IEEEkeywords}

\section{Introduction}
\label{sec:intro}
Image denoising is a foundational task of low-level vision, which aims to remove unwanted noise signals from input images and recover noise-free clean images. The task is widely used as a pre-processing technique for many high-level vision applications, such as image classification, object detection, and semantic segmentation.

Traditional methods leverage image priors to solve the degradation problem~\cite{elad2006image,rudin1992nonlinear}. However, these methods are highly dependent on hand-crafted features and are time-consuming, which limits applications in real-world scenarios. Recently, CNN-based denoising methods~\cite{zhang2017beyond,zamir2021multi} have achieved remarkable performance. Although convolutional operation provides inductive bias of local connectivity and translation equivariance, the limited receptive field hinders modeling long-range dependencies over a whole image.

\begin{figure}[t]
    \centering
    \includegraphics[width=.9\linewidth]{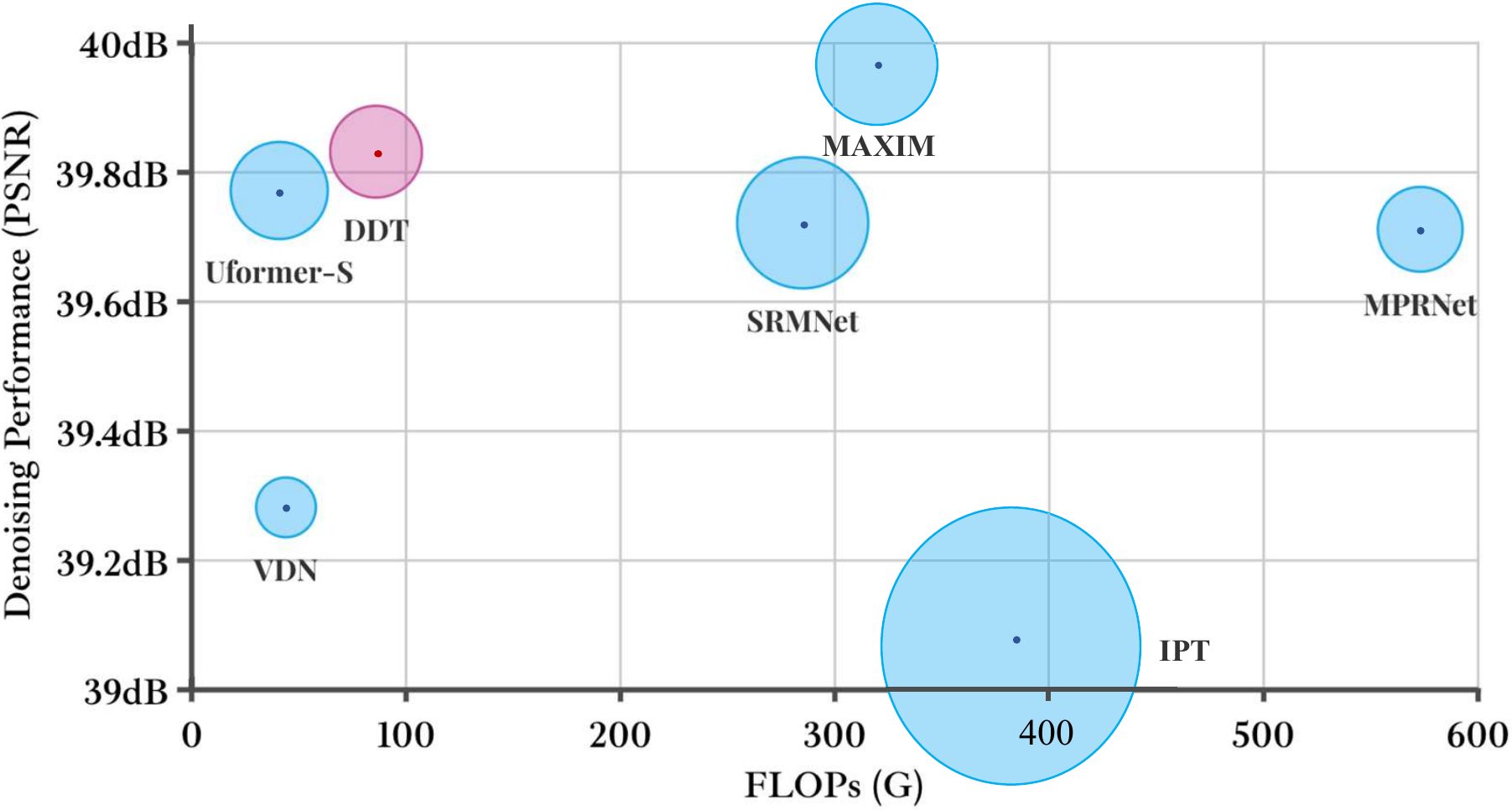}
    \vspace{-0.1in}
    \caption{Denoising results (PSNR) $vs.$ computational cost (FLOPs) on SIDD dataset~\cite{abdelhamed2018high}. The radius of circle represents the number of parameters.}
    \label{fig:fig1}
  \end{figure}

Inspired by the astounding performance of Transformer models in NLP field~\cite{vaswani2017attention}, research has moved towards applying the same principles in computer vision~\cite{dosovitskiy2020image,liu2021swin,wang2021pyramid}. The feature representations generated from self-attention components do not contain the spatial constraints imposed by convolutional operations. However, the standard self-attention suffers from quadratic complexity with the length of sequences, making it hard to apply to the pixel level directly.

To solve these problems, some low-level Transformer models~\cite{wang2022uformer,liang2021swinir,zamir2022restormer} utilize local or channel-wise attention but not fully exploiting the global modeling advantage of the Transformer. We consider local and global information equally useful for image denoising, so we use a parallel structure to process both of them efficiently. Besides, highly similar patterns often appear repeatedly, and it is redundant to calculate all these regions. Therefore, we propose an adaptively deformable operation to reduce these redundant calculations while focusing on more informative areas.

\begin{figure*}[t]
  \centering
  \includegraphics[width=.95\linewidth]{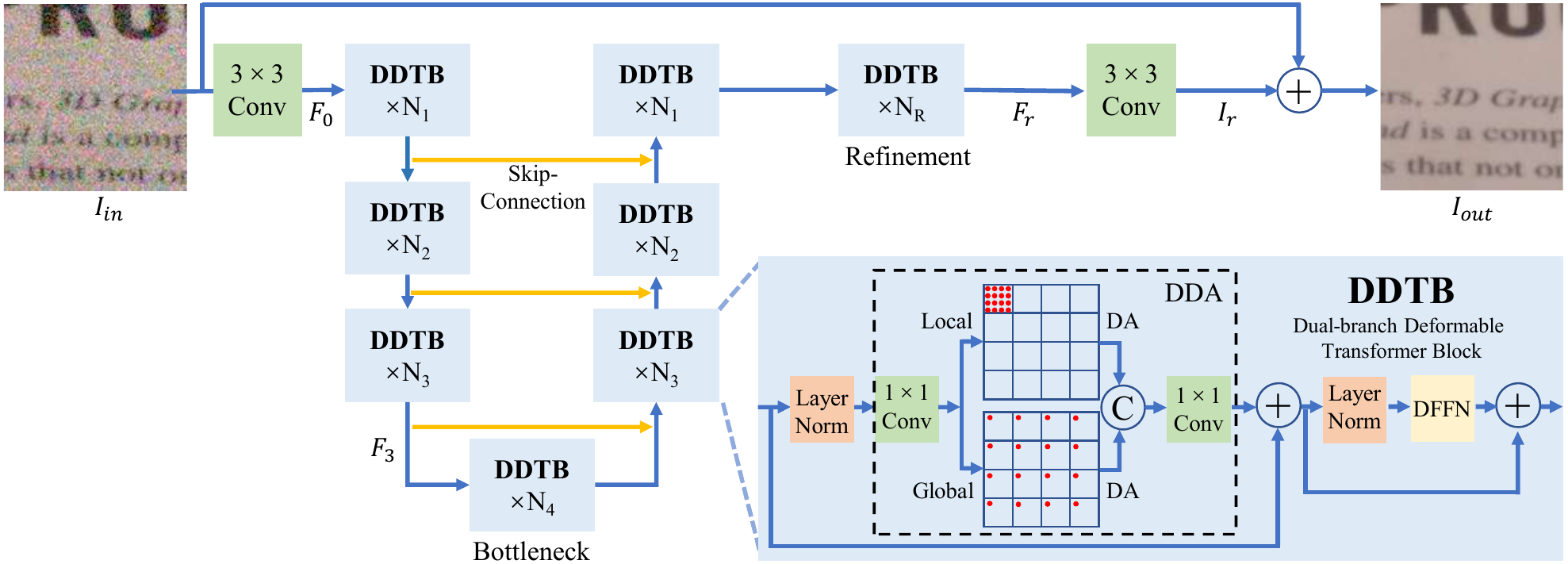}
  \caption{The overall architecture of the Dual-branch Deformable Transformer. The Unet-like hierarchical architecture with an extra refinement stage is used for image denoising, and each stage contains a different number of Dual-branch Deformable Transformer Blocks (DDTBs). The DDTB captures the local and global features parallelly using Dual-branch Deformable Attention (DDA) and Depth-wise Feed-Forward Network (DFFN).}
  \label{fig:pipline}
\end{figure*}

In this paper, we propose an effective Dual-branch Deformable Transformer (\system) network for image denoising. \system applies parallelly local and global branches with the deformable attention strategy. Specifically, we first divide the input feature into patches using different rules in both branches, then perform deformable attention inside patches in the local branch and among the same relative locations of each patch in the global branch. In the deformable attention, we reduce the number of key-value pairs, preserving more informative ones adaptively and saving calculational costs furtherly. In this way, global receptive fields and local aggregations are both involved in our Transformer model with linear complexity by the dual-branch structure, and the deformable attention mechanism also reduces redundant calculations.
The main contributions of our work are summarized as follows:
\begin{itemize}
\item We propose an efficient Dual-branch Deformable Transformer denoising network that can process high-resolution noisy images in linear complexity.
\item Deformable attention is introduced to reduce redundant calculations and focus on more informative regions.
\item Extensive experiments prove that our method can produce more competitive performance with fewer computational costs than other state-of-the-art methods.
\end{itemize}

\section{Related Work}
\label{sec:related}
\subsection{Image Denoising}
Image denoising, as a fundamental low-level task, has been studied for decades. Traditional methods usually leverage image priors and handcrafted features. The representative block-matching 3D (BM3D)~\cite{dabov2007image} utilizes the nonlocal similarity and transform-domain technique and follows three consecutive stages: grouping, collaborative filtering, and aggregation.

With the development of deep learning, many learning-based denoising methods have been proposed and achieved remarkable performance. GCBD~\cite{chen2018image} introduces GAN~\cite{radford2015unsupervised} to image denoising tasks. GCDN~\cite{valsesia2020deep} captures self-similar information for denoising. FFDNet~\cite{zhang2018ffdnet} takes a tunable noise level map as input and enhances flexibility to deal with spatially variant noise. MPRNet~\cite{zamir2021multi} uses a multi-stage architecture to achieve state-of-the-art. Besides, some works try to extract global features to improve denoising results. MAXIM~\cite{tu2022maxim} explores the potential of MLP in image restoration, and others~\cite{zamir2022restormer,wang2022uformer,chen2021ipt,liang2021swinir} utilize attention mechanisms~\cite{vaswani2017attention} for global modeling. However, these methods have characteristics of high memory occupation and time consumption.

\subsection{Vision Transformer}
Transformer structure was first proposed for natural language processing tasks~\cite{vaswani2017attention} with multi-head self-attention. Recently, the Vision Transformer (ViT)~\cite{dosovitskiy2020image} has surpassed previous state-of-the-art CNN-based models on the image classification task by projecting image patches into sequences and feeding them to Transformer blocks. PVT~\cite{wang2021pyramid} uses a pyramid architecture to leverage multi-scale features and improves performance on object detection and instance segmentation tasks. Swin-Transformer~\cite{liu2021swin} proposes shifted window-based attention mechanism, reducing the computational cost compared with ViT. Reference \cite{zhu2020deformable,xia2022vision} introduce a deformable strategy to reduce the number of key-value pairs in Transformer.

Inspired by the long-range dependencies modeling ability of Vision Transformers, researchers also exhibit its potential in low-level tasks. IPT~\cite{chen2021ipt} proposes a pre-trained backbone for image restoration. In order to reduce high computational complexity, SwinIR~\cite{liang2021swinir} and Uformer~\cite{wang2022uformer} apply attention in local windows with shifted window operation. Restormer~\cite{zamir2022restormer} leverages attention on channel dimension as a global operation, avoiding huge calculation costs on spatial level but lacking locality. Different from previous works, our novel framework DDT can leverage both local and global contexts effectively and demonstrates high efficiency.

\begin{figure*}[t]
    \centering
    \includegraphics[width=.95\linewidth]{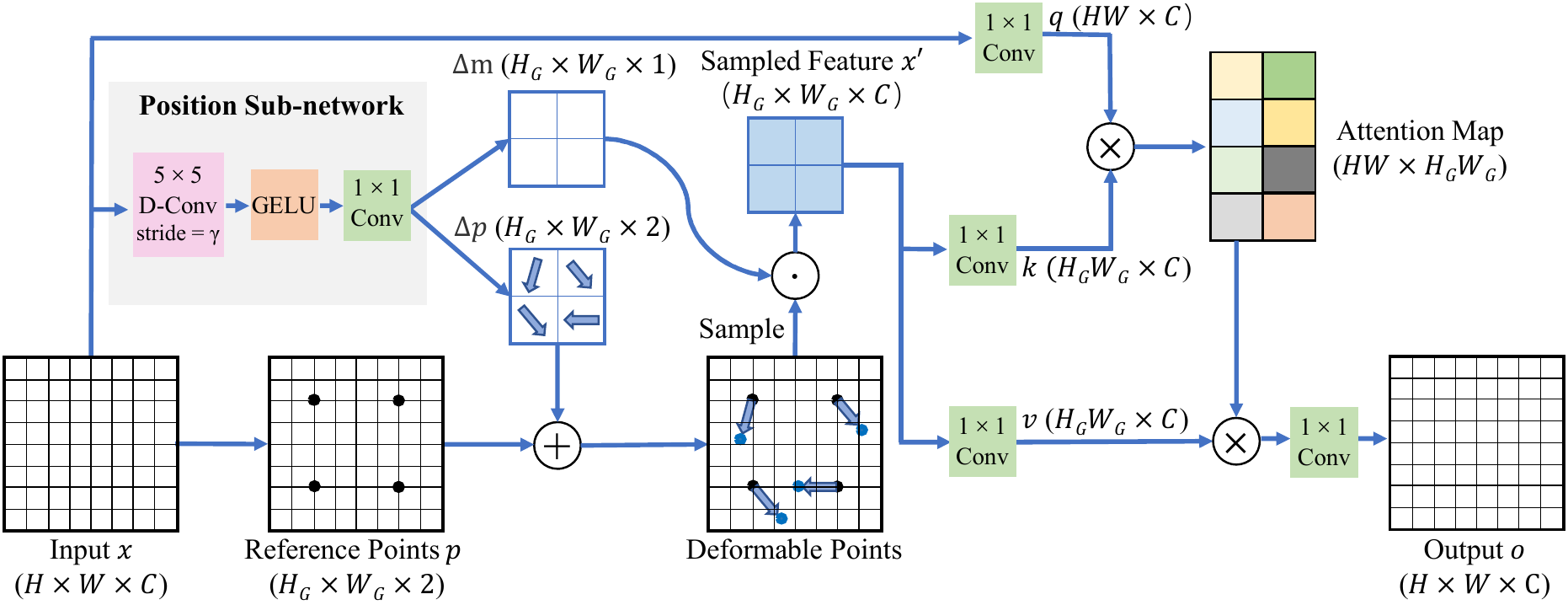}
    \caption{The structure of deformable attention. A group of offsets $\Delta p$ are learned from the input $x$ and added on the pre-defined reference points $p$ to get deformable points. Features $x'$ sampled from the input at the positions of deformable points are then projected as keys($k$) and values($v$), while queries($q$) are still from the input $x$. 
    }
    \label{fig:DA}
\end{figure*}

\section{Approach}
\label{sec:method}
Our main purpose is to develop an efficient Transformer model that can process high-resolution images the image denoising tasks. In order to enhance feature representations, we use a dual-branch structure to parallelly capture local and global features. In addition, we use a deformable strategy to focus attention on more important regions and further reduce computational costs. The overall pipeline of the Dual-branch Deformable Transformer is described as Fig.~\ref{fig:pipline}.

\subsection{Overall Pipeline}
\label{subsec:overall}
Given a noisy image $I_{in}\in\mathbb{R}^{H\times W \times 3}$, the input is first projected into the feature $F_0\in\mathbb{R}^{H\times W \times C}$ by a 3$\times$3 convolution layer.
Then the feature $F_0$ goes through a 4-stage Unet-like encoder-decoder architecture. Each stage includes multiple DDTBs and a sample layer (\ie~downsample or upsample).
The output feature $F_3\in\mathbb{R}^{\frac{H}{8} \times \frac{W}{8} \times 8C}$ goes through the Bottleneck and is fed into the decoder.The decoder has a symmetrical structure, where an upsample layer is used at the beginning of each stage. The skip-connection~\cite{ronneberger2015u} is used to boost performance, and we use the concatenation operation and a 1$\times$1 convolution layer for feature fusion. Inspired by~\cite{zamir2022restormer}, we introduce a refinement stage after the decoder, which aims to enhance feature representation for more details of images. Finally, the feature $F_r$ from the refinement stage through a 3$\times$3 convolution filter to obtain the residual image $I_r\in\mathbb{R}^{H\times W \times 3}$, and the restored output $I_{out}$ is denoted as:
\begin{equation}
    I_{out} = I_{in} + I_r
\end{equation}

\subsection{Dual-branch Deformable Transformer Block}
\label{subsec:block}
DDTB consists of two main components: Dual-branch Deformable Attention (DDA) and Depth Feed-Forward Network (DFFN). A layer normalization is first applied in each part. Specifically, DDA splits the feature over the channel dimension and sends them into a dual-branch structure, where local and global interactions are performed parallelly. In the local branch, we divide the feature into non-overlapping patches with pre-defined patch sizes and apply the spatial attention mechanism inside the patches. In the global brunch, we use a pre-defined number of patches to do the patch partitioning and perform calculations among the corresponding positions of each patch. The multi-scale features output by the dual-branch are then fused by concatenation operation and a linear layer. We utilize a deformable attention mechanism as the spatial operation in both branches, which will be introduced in Section~\ref{subsec:DA}.

Inspired by~\cite{zamir2022restormer, lee2022knn}, we add depth-wise convolution into the standard Feed-Forward Network~\cite{dosovitskiy2020image} as DFFN to further enhance features. Formally, The $l$-th DDTB is formulated as:
\begin{equation}
    \hat{X_l}=\texttt{DDA}(\texttt{LN}(X_{l-1}))+X_{l-1}
\end{equation}
\begin{equation}
    X_l=\texttt{DFFN}(\texttt{LN}(\hat{X_l}))+\hat{X_l}
\end{equation}
where $X_{l-1}$ represents the output of $(l-1)th$ DDTB, and \texttt{LN} means the layer normalization. The residual connection~\cite{he2016deep} is used in both components.
With the dual-branch structure, we can process inputs with linear complexity, making it possible to process high-resolution images using Transformer architecture, preserving global receptive fields while capturing locality.

\subsection{Deformable Attention}
\label{subsec:DA}
To further save computational costs, the deformable attention (Fig.~\ref{fig:DA}) is applied for efficient spatial operation. Inspired by~\cite{xia2022vision}, we first define a set of grid-shape reference points $p\in\mathbb{R}^{H_G\times W_G \times 2}$ based on the input $x\in\mathbb{R}^{H\times W \times C}$. The number of points is determined by a hyper-parameter $\gamma$, $H_G=H/\gamma$ and $W_G=W/\gamma$. Meanwhile, we feed  $x$ into a Position Sub-network containing a $5\times 5$ depth-wise convolutional layer with stride $\gamma$. The output of the sub-network has three channels, and the first two channels are deformable offsets ${\Delta} p\in\mathbb{R}^{H_G\times W_G \times 2}$ for each reference point. We add the offsets $\Delta p$ on each point to obtain the deformable points and sample the feature $x'\in\mathbb{R}^{H_G\times W_G \times C}$ from $x$. The remaining channel is used as a modulation scalar ${\Delta}m\in\mathbb{R}^{H_G\times W_G \times 1}$ to multiply $x'$, aiming to enhance the spatial representations.
The sampled feature $x'$ can be expressed as:
\begin{equation}
    x'=\psi (x,p+\Delta p)\odot \Delta m
\end{equation}
where $\odot$ denotes element-wise multiplication, and $\psi(\cdot ,\cdot )$ represents the sampling process. We leverage bilinear interpolation to make this process differentiable.

Then we apply linear projections on $x$ and $x'$ to obtain queries($q$), keys($k$) and values($v$) and perform multi-head attention with $M$ heads to get the output $o$:
\begin{equation}
    q = W_{q}x, k=W_{k}x', v=W_{v}x'
\end{equation}
\begin{equation}
    z^{(m)}=\texttt{softmax}(\frac {q^{(m)}k^{(m)\top}} {\sqrt{d}})v^{(m)}, m=1,2,...,M
\end{equation}
\begin{equation}
    o = W_{o}\texttt{concat}(z^{(1)}, z^{(2)},..., z^{(m)})
\end{equation}
where $W_{(\cdot )}\in\mathbb{R}^{C\times C}$ is the $1{\times}1$ convolution for linear projection, and $d=C/M$ is the number of channels for each head. $z^{(m)}, q^{(m)}, k^{(m)}, v^{(m)}$ represent the output, queries, keys and values in the $m$-th head. It is noteworthy that $k$ and $v$ are generated from the sampled feature $x'$ with size $H_{G}W_{G}\times C$, which is smaller than $q$.

We consider that the sampled feature $x'$ preserves most of the important information of $x$, reducing redundant contents compared with the origin feature $x$. It is a data-dependent way that can adapt to different inputs. In this way, we reduce the computation on redundant areas and let the model focus on more informative regions.

\subsection{Analysis of Computational Cost}
Our method consists of DDTBs with different scales. The computational costs of DDTB come mainly from Dual-branch Deformable Attention (DDA), which consists of two convolutional layers and a dual-branch structure. We assume the input size for DDA as $H\times W\times C$, and the patch size and the number of patch in the local and global branches as $p \times p$ equally.

\vspace{0.08in}
\noindent\textbf{Convolutional layers.}
There are two 1$\times$1 convolutional layers in DDA for channel expanding and feature fusion, respectively. We denote each of them as:
\begin{equation}
    \Omega(\text{Conv})=2HWC^2
\end{equation}

\vspace{0.08in}
\noindent\textbf{Dual-branch structure.}
The deformable attention (DA) is applied in the local and global branches. We take the local branch as an example to illustrate the computational costs. We first divide the input into patches with size $p\times p\times C$, so we get $HW/p^{2}$ patches. We feed them into DA for local feature extraction. The hyper-parameter in DA for controling the number of keys and values is $\gamma$. We take each branch's costs as:
\begin{equation}\label{eq:branch}
        \Omega(\text{Branch})=\frac{2{\gamma}^2+2}{{\gamma}^2}HWC^2+\frac{2p^2+29}{{\gamma}^2}HWC+\frac{2}{{\gamma}^2}HW
\end{equation}
The global branch divides the input into $p\times p$ patches and   perform DA among the corresponding positions of each patch, the costs of which are same as equation (\ref{eq:branch}).

Overall, we present the computational costs of DDA as:
\begin{equation}
    \begin{split}
    \Omega(\text{DDA})&=2\Omega(\text{Conv})+2\Omega(\text{Branch})\\&=\frac{8{\gamma}^2+4}{{\gamma}^2}HWC^2+\frac{4p^2+58}{{\gamma}^2}HWC+\frac{4}{{\gamma}^2}HW
\end{split}
\end{equation}
where the computational costs are linear with the spatial resolution $HW$, indicating that it is possible to apply the proposed DDT on high-resolution images frequently appearing in image denoising tasks. 

\begin{figure*}[t]
    \centering
    \includegraphics[width=.99\linewidth]{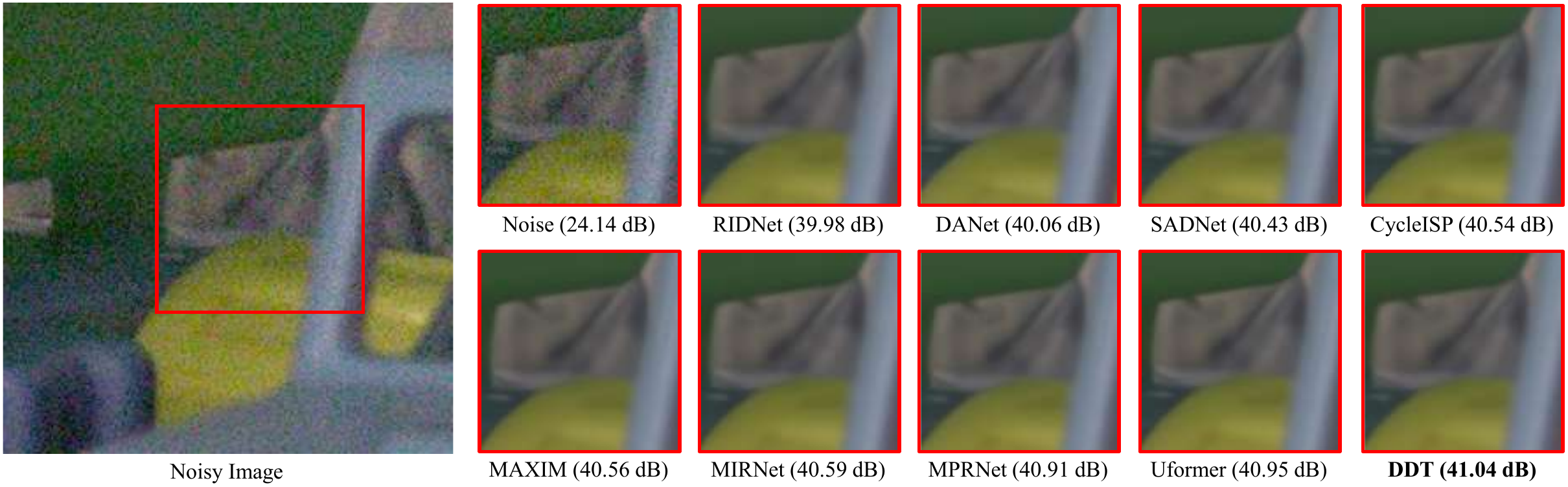}
    \caption{Visual comparisons with state-of-the-art methods on real-world image denoising.}
    \label{fig:denoising}
\end{figure*}
    
\section{Experiments}
\label{sec:exp}
In this section, we first introduce datasets and implementation details. Next, we describe extensive comparisons with previous methods. Moreover, we perform ablation studies to evaluate the effectiveness of each component.
    
\subsection{Datasets}
\noindent\textbf{Real-world denoising.} For real-world denoising, we train \system on the SIDD~\cite{abdelhamed2018high} training set and test on both SIDD validation set and DND online benchmark~\cite{plotz2017benchmarking}. SIDD provides noisy-clean image pairs from 10 scenes under different lighting conditions using five representative smartphone cameras. DND is an online benchmark of 50 image pairs captured with four consumer cameras.

\vspace{0.06in}
\noindent\textbf{Synthetic denoising.} We add Gaussian noise to some natural image datasets~\cite{agustsson2017ntire, timofte2017ntire} with random sigma ranging from [0, 50] to generate noisy-clean image pairs for training. We select Gaussian noise with sigma 15, 25, and 50 on datasets CBSD68~\cite{martin2001database}, Kodak24~\cite{kodak}, McMaster~\cite{zhang2011color} and Urban100~\cite{huang2015single} as testsets.
  
\subsection{Implementation Details}
  \begin{table}[t]
    \footnotesize{
    \centering
    \caption{The quantitative results of real-world denoising on SIDD and DND datasets. The bold numbers represent the best results. \\F and P denote FLOPs and Parameters, respectively.}
      \begin{tabular}{l|c|c|c|c|c|c}
      \hline
      \multicolumn{1}{c|}{\multirow{2}[0]{*}{\textbf{Method}}} & \multicolumn{1}{c|}{\multirow{2}[0]{*}{\textbf{F(G)}}} & \multirow{2}[0]{*}{\textbf{P(M)}} & \multicolumn{2}{c|}{\textbf{SIDD~\cite{abdelhamed2018high}}} & \multicolumn{2}{c}{\textbf{DND~\cite{plotz2017benchmarking}}} \\
      \cline{4-7} & & & \textbf{PSNR}  & \textbf{SSIM}  & \textbf{PSNR}  & \textbf{SSIM} \\ \hline
      Restormer~\cite{zamir2022restormer} & 155 & 26.1 & \textbf{40.02} & \textbf{0.960} & 40.03 & \textbf{0.956}\\
      MAXIM~\cite{tu2022maxim} & 339 & 22.2 & 39.96 & \textbf{0.960} & 39.84 & 0.954 \\
      Uformer-B~\cite{wang2022uformer} & 86 & 50.9 & 39.89 & \textbf{0.960} & \textbf{40.04} & 0.956 \\
      SRMNet~\cite{fan2022selective} & 285 & 37.6 & 39.72 & 0.959 & 39.44 & 0.951 \\
      MIRNet~\cite{zamir2020learning} & 785 & 31.8 & 39.72 & 0.959 & 39.88 & \textbf{0.956} \\
      MPRNet~\cite{zamir2021multi} & 573 & 15.7 & 39.71 & 0.958 & 39.80 & 0.954 \\
      CycleISP*~\cite{zamir2020cycleisp} & 184 & 2.8 & 39.52 & 0.957  & 39.56 & \textbf{0.956}\\
      DANet~\cite{yue2020dual} & 30 & 63.0 & 39.30 & 0.916 & 39.59 & 0.955 \\
      VDN~\cite{yue2019variational} & 44 & 7.8 & 39.28 & 0.909 & 39.38 & 0.952 \\
      IPT*~\cite{chen2021ipt} & 380 & 115.3 & 39.10 & 0.954 & 39.62 & 0.952 \\ 
      RIDNet*~\cite{anwar2019real} & 98 & 1.5 & 38.71 & 0.914 & 39.26 & 0.953 \\ \hline
      \system (ours) & 86 & 18.4 & 39.83 & \textbf{0.960} & 39.78 & 0.954 \\
      \hline
      \multicolumn{7}{l}{* denotes methods using additional training data.}
      \end{tabular}
      \label{table:SIDDexperiment}
    }
  \end{table}

  \begin{table*}[t]
    \centering
    \small
    \caption{Comparison results of synthetic denoising. The \textbf{bold} and \underline{underline} represent the best and the secend best results.}
      \begin{tabular}{l|c|c|ccc|ccc|ccc|ccc}
      \hline
      \multirow{2}{*}{\textbf{Method}} &
      \multirow{2}{*}{\textbf{F(G)}} &
      \multirow{2}{*}{\textbf{P(M)}} &
      \multicolumn{3}{c|}{\textbf{CBSD68~\cite{martin2001database}}} &
      \multicolumn{3}{c|}{\textbf{Kodak24~\cite{kodak}}} &
      \multicolumn{3}{c|}{\textbf{McMaster~\cite{zhang2011color}}} &
      \multicolumn{3}{c}{\textbf{Urban100~\cite{huang2015single}}} \\ \cline{4-15} 
      &  &  & \textbf{$\sigma$=15}     & \textbf{$\sigma$=25}     & \textbf{$\sigma$=50} & \textbf{$\sigma$=15} & \textbf{$\sigma$=25} & \textbf{$\sigma$=50} & \textbf{$\sigma$=15} & \textbf{$\sigma$=25} & \textbf{$\sigma$=50} & \textbf{$\sigma$=15} & \textbf{$\sigma$=25} & \textbf{$\sigma$=50}\\ \hline
      IRCNN~\cite{zhang2017learning} &49  &0.7   &33.86 &31.16 &27.86 &34.69 &32.18 &28.93 &34.58 &32.18 &28.91 &33.78 &31.20 &27.70 \\
      DnCNN~\cite{zhang2017beyond} &93  & 1.4  &33.94 &31.30 &28.02 &34.64 &32.14 &28.99 &34.69 &32.32 &38.23 &33.80 &31.34 &27.98 \\
      DSNet~\cite{peng2019dilated} &- &-   &33.91 &31.28 &28.05 &34.63 &32.16 &29.05 &34.67 &32.40 &29.28 &- &- &- \\
      MLEFGN~\cite{fang2020multilevel} &447  &6.9  &- &- &28.21 &- &- &29.38 &- &- &- &- &- &28.92 \\
      IPT~\cite{chen2021ipt} &380  &115.3  &- &- &28.39 &- &- &29.64 &- &- &29.98 &- &- &\underline{29.71} \\
      Restormer~\cite{zamir2022restormer} &155  &26.1  &\textbf{34.39} &\textbf{31.78} &\textbf{28.59} &\textbf{35.44} &\textbf{33.02} &\textbf{30.00} &\textbf{35.55} &\textbf{33.31} &\textbf{30.29} &\textbf{35.06} &\textbf{32.91} &\textbf{30.02} \\ \hline
      DDT (ours) &86  &18.4   &\underline{34.30} &\underline{31.69} &\underline{28.50} &\underline{35.31} &\underline{32.88} &\underline{29.85} &\underline{35.36} &\underline{33.11} &\underline{30.07} &\underline{34.80} &\underline{32.58} &29.56 \\
      \hline
      \end{tabular}
      \label{table:GaussianResult}
  \end{table*}\label{ss4.2}
\system adopts a 4-stage encoder-decoder architecture. Following~\cite{zamir2022restormer}, we set the number of DDTBs from the 1st stage to Bottleneck as (4, 6, 6, 8) with the number of attention heads (1, 2, 4, 8) and 4 extra blocks for the Refinement stage. The channel number in the 1st stage is set as 32, which will increase gradually in deeper stages. We use AdamW~\cite{loshchilov2017decoupled} optimizer(${\beta}_1=0.9$, ${\beta}_2=0.99$, weight decay $1e^{-4}$) and $L_1$ loss with 300K iterations for optimization. The learning rate is initialized as $3e^{-4}$ and reduce to $1e^{-6}$ with the cosine annealing scheduler~\cite{loshchilov2016sgdr}. Progressive learning strategy~\cite{zamir2022restormer} from $128 \times 128$ to $256 \times 256$ is also used, and we utilize rotation and flips for data augmentations. All training processes are conducted on 4 NVIDIA TITAN Xp GPUs.

\subsection{Comparison with State-of-the-arts}
\noindent\textbf{Real-world denoising.}
Table~\ref{table:SIDDexperiment} reports the real-world denoising results on the SIDD and DND datasets. We compare \system with state-of-the-art methods, including CNN-based, MLP-based and Transformer-based methods. Our proposed \system achieves competitive performance on the SIDD dataset, getting state-of-the-art results on SSIM with the fewest parameters and FLOPs. MAXIM~\cite{tu2022maxim} and Restormer~\cite{zamir2022restormer} surpass our DDT by 0.13dB and 0.19dB on PSNR, but ours only takes 25.4$\%$ and 55.5$\%$ FLOPs compared to them respectively, demonstrating the efficiency of \system. Uformer-B uses 50.9M parameters and gets better results, but our DDT performs better when Uformer-S uses a similar number of parameters. To show results more clearly, we present the PSNR vs. FLOPs in Fig.~\ref{fig:fig1}, in which the radius of each point represents the number of parameters. We also show the visual results with various state-of-the-art methods on SIDD in Fig.~\ref{fig:denoising}, and it is obviously observed that \system not only removes noise but preserves sharp and clear details.
  
\vspace{0.06in}
\noindent\textbf{Synthetic denoising.}
Table~\ref{table:GaussianResult} shows PSNR values of different methods for synthetic denoising. We select various methods using consistent experiment settings for comparison, including IRCNN~\cite{zhang2017learning}, DnCNN~\cite{zhang2017beyond}, DSNet~\cite{peng2019dilated}, MLEFGN~\cite{fang2020multilevel}, IPT~\cite{chen2021ipt} and Restormer~\cite{zamir2022restormer}. DDT efficiently achieves suboptimal performance only lower than Restormer, but Restormer takes 1.8x more computational costs. Our results surpass another Transformer model IPT on CBSD68, Kodak24 and McMaster datasets, which uses 6.26$\times$ more parameters and 4.42$\times$ more FLOPs compared with our DDT.
  
\begin{table}[t]
    \centering
    \caption{Effects of dual-branch structure.}
    \label{table:ablation1}

    \setlength{\tabcolsep}{4.6mm}
    \begin{tabular}{cc|c|c|c}
    \hline
    \textbf{Local} & \textbf{Global} & \textbf{PSNR} & \textbf{F (G)} & \textbf{P (M)} \\ \hline
    $\checkmark$ & & 39.77 & 86 & 18.4 \\
          & $\checkmark$ & 39.78 & 86 & 18.4 \\
          $\checkmark$ & $\checkmark$ & \textbf{39.83} & 86 & 18.4 \\ \hline
    \end{tabular}
\end{table}
  
\begin{table}[t]
    \centering
    \caption{Comparison with different spatial operations. }
    \label{table:ablation2}
    \setlength{\tabcolsep}{6mm}
    \begin{tabular}{l|c|c|c}
    \hline
    \textbf{Operation} & \textbf{PSNR} & \textbf{F (G)} & \textbf{P (M)} \\ \hline
     MLP & 39.65 & 121 & 18.4 \\
     MHSA & 39.76 & 157 & 18.3 \\
     DA (ours) & \textbf{39.83} & 86 & 18.4 \\ \hline
    \end{tabular}
\end{table}
  
\subsection{Ablation Studies}
In this section, we conduct ablation studies to evaluate the performance of the dual-branch structure and deformable attention on the SIDD dataset.
  
\vspace{0.06in}
\noindent
\textbf{Dual-branch.}
To evaluate the effectiveness of dual-branch structure, we conduct experiments with a single branch (\ie~only using local or global branch). For a fair comparison, we double the number of channels when using a single branch to keep the same parameters as the dual-branch's. The results are shown in Table~\ref{table:ablation1}. We observed similar results when using a single branch. However, by combining them into the parallelly dual-branch structure, the PSNR improved by 0.06dB and 0.05dB, respectively.
  
\vspace{0.06in}
\noindent
\textbf{Deformable attention.}
Deformable attention reduces redundant computations and focuses on more important regions. We compared with multilayer perceptron (MLP) and standard multi-head self-attention (MHSA)~\cite{dosovitskiy2020image} which have quadratic complexity to show our deformable attention's effectiveness and high efficiency. We adjust the number of layers in MLP and the number of channels in MHSA to maintain a similar number of parameters ($\sim$18.4M). Table~\ref{table:ablation2} shows the detailed results that our deformable attention achieves the best performance with only 71.1$\%$ and 54.8$\%$ computational costs of MLP and MHSA.

\section{Conclusions}
\label{sec:conclusion}
In this paper, we present an efficient and effective Transformer model DDT for image denoising, which is computationally efficient for processing high-resolution images.
Specifically, we use a dual-branch structure in Dual-branch Deformable Transformer Blocks, parallel modeling local and global information with linear complexity.
To further save computational resources, deformable attention is also applied, reducing redundant calculations and focusing on more informative regions.
Experimental results on synthetic and real-world denoising tasks demonstrate that our method achieves competitive results using a small number of computations.

\bibliographystyle{IEEEbib}
\bibliography{total}

\newpage

\appendix
In this appendix, We will show the derivation process of the equations in Section 3.4.
\subsection{Deformable Attention}
Assuming the input feature size to the deformable attention (DA) is $H\times W\times C$, and the stride in DA's Position Sub-network is $\gamma$. We divide the computations of DA into four parts: Position Sub-network, Convolutional layers, $\Delta p$ and $\Delta m$, and multi-head attention.

\vspace{0.08in}
\noindent\textbf{Position sub-network.}
There is a 5$\times$5 depth-wise convolutional layer with stride $\gamma$ and a 1$\times$1 convolutional layer to reduce the number of channel to 3. Therefore, we can summarize the Position Sub-network (PS)'s computational costs as:
\begin{equation}
    \label{eq:PS}
    \Omega (\text{PS}) = 5^2 \cdot \frac{HW}{r^2}C + 3 \cdot \frac{HW}{r^2}C
\end{equation}

\vspace{0.08in}
\noindent\textbf{Convolutional layers.}
There are four 1$\times$1 convolutional layers for generating $q$, $k$, $v$ and $o$. Based on the different input sizes, we denote their costs as:
\begin{equation}
    \Omega (\text{q})=HW{C^2}
\end{equation}

\begin{equation}
    \Omega (\text{k})=\frac{HW}{r^2}C^2
\end{equation}

\begin{equation}
    \Omega (\text{v})=\frac{HW}{r^2}C^2
\end{equation}

\begin{equation}
    \Omega (\text{o})=HW{C^2}
\end{equation}

\vspace{0.08in}
\noindent\textbf{$\Delta $p and $\Delta $m.}
The $\Delta p$ and $\Delta m$ output by the Position Sub-network are then element-wise added on the reference points and element-wise multiplied on the sampled feature, respectively. We express their costs as:
\begin{equation}
    \Omega(\text{$\Delta p$})=2\cdot\frac{HW}{r^2}
\end{equation}

\begin{equation}
    \Omega(\text{$\Delta m$})=\frac{HW}{r^2}C
\end{equation}

\vspace{0.08in}
\noindent\textbf{Multi-head attention.}
After obtaining $q$, $k$ and $v$, Multi-head attention (MHA) is applied. there existing 2 matrix multiplications among $q$, $k$, $v$.
\begin{equation}
    \label{eq:MHA}
    \Omega(\text{MHA})=2\cdot\frac{H^2 W^2}{r^2}C
\end{equation}

We summarize the overall computational costs of the deformable attention (DA)  by adding up Equation (\ref{eq:PS}) - (\ref{eq:MHA}) as:
\begin{equation}\label{eq:DA}
    \begin{split}
        \Omega(\text{DA})=&\frac{2}{{\gamma}^2}H^2W^2C+(2+\frac{2}{{\gamma}^2})HWC^2\\&+\frac{29}{{\gamma}^2}HWC+\frac{2}{{\gamma}^2}HW
    \end{split}
\end{equation}

\subsection{Dual-branch Deformable Attention}
Dual-branch Deformable Attention (DDA) consists of two 1$\times$1 convolutional layers and a dual-branch structure.
We assume the input size for DDA as $H\times W\times C$, and the patch size and the number of patch in the local and global branches as $p \times p$ equally.

\vspace{0.08in}
\noindent\textbf{Convolutional layers.}
There are two 1$\times$1 convolutional layers in DDA. One is used for expanding channels before the dual-branch structure, and the other for multi-scale features fusion after the dual-branch. We denote each of them as:
\begin{equation}
    \Omega(\text{Conv})=2HWC^2
\end{equation}

\vspace{0.08in}
\noindent\textbf{Local and global branch.}
The DA is applied inside and among the patches in the local and global branches, respectively. We take the local branch as an example to illustrate the computational costs. We first divide the input into patches with size $p\times p\times C$, so we get $HW/p^{2}$ patches. We feed them into DA for local feature extraction. We take the patch size into equation (\ref{eq:DA}) to get one branch's costs as:
\begin{equation}\label{eq:branch_app}
    \begin{split}
        &~~~~~\Omega(\text{Branch})\\&=\frac{HW}{p^2}(\frac{2}{{\gamma}^2}p^4C+(2+\frac{2}{{\gamma}^2})p^2C^2+\frac{29}{{\gamma}^2}p^2C+\frac{2}{{\gamma}^2}p^2)\\&=\frac{2{\gamma}^2+2}{{\gamma}^2}HWC^2+\frac{2p^2+29}{{\gamma}^2}HWC+\frac{2}{{\gamma}^2}HW
    \end{split}
\end{equation}
The global branch divides the input into $p\times p$ patches and applied DA among patches, the costs of which are same as equation (\ref{eq:branch_app}).

Finally, we present the computational costs of DDA as:
\begin{equation}
    \begin{split}
    \Omega(\text{DDA})&=2\Omega(\text{Branch})+2\Omega(\text{Conv})\\&=\frac{8{\gamma}^2+4}{{\gamma}^2}HWC^2+\frac{4p^2+58}{{\gamma}^2}HWC+\frac{4}{{\gamma}^2}HW
\end{split}
\end{equation}

\end{document}